\documentclass[12pt]{iopart}

\usepackage{graphicx}
\usepackage{fancyhdr}
\usepackage{amssymb}
\usepackage{bm}
\usepackage{hyperref}
\usepackage{xcolor}

\begin{document}
\pagestyle{fancy}

\title{Known Operator Learning and Hybrid Machine Learning in Medical Imaging --- A Review of the Past, the Present, and the Future}
\lhead{Known Operator Learning and Hybrid Machine Learning}
\author{Andreas Maier, Harald Köstler, Marco Heisig, Patrick Krauss, Seung Hee Yang}

\address{Friedrich-Alexander-University Erlangen-Nuremberg, Germany}
\ead{andreas.maier@fau.de}
%\vspace{10pt}
\begin{indented}
\item[]July 2021
\end{indented}

\begin{abstract}
In this article, we perform a review of the state-of-the-art of hybrid machine learning in medical imaging. We start with a short summary of the general developments of the past in machine learning and how general and specialized approaches have been in competition in the past decades. A particular focus will be the theoretical and experimental evidence pro and contra hybrid modelling. Next, we inspect several new developments regarding hybrid machine learning with a particular focus on so-called known operator learning and how hybrid approaches gain more and more momentum across essentially all applications in medical imaging and medical image analysis. As we will point out by numerous examples, hybrid models are taking over in image reconstruction and analysis. Even domains such as physical simulation and scanner and acquisition design are being addressed using machine learning  grey box modelling approaches. Towards the end of the article, we will investigate a few future directions and point out relevant areas in which hybrid modelling, meta learning, and other domains will likely be able to drive the state-of-the-art ahead.
\end{abstract}

%\maketitle

\newcommand{\fex}{\bm x}
\newcommand{\param}{\bm \theta}
\newcommand{\activation}{h}
\newcommand{\real}{{\rm I\!R}}
\newcommand{\prediction}{\hat{y}}
\newcommand{\classvar}{y}
\newcommand{\classifier}{\hat{f}}
\newcommand{\loss}{L}
\newcommand{\avgerror}{\bar{\mathcal{E}}}

\section{Introduction}

In the past decade, machine learning and in particular deep learning \cite{goodfellow2016deep} have made a tremendous impact on the field of medical imaging. We have seen impact on all areas of the field ranging from the acquisition side all the way to the image analysis~\cite{maier2019gentle}. Recently, more and more research explores the concept of hybrid machine learning approaches \cite{wurfl2016deep, kobler2017variational, nguyen2018rendernet, maier2018precision, meister2018towards, maier2019learning, wickramasinghe2020voxel2mesh}. The key idea is to re-use classical methods ranging from geometrical modelling such as projection \cite{wurfl2016deep} and rendering \cite{nguyen2018rendernet} to the full integration of mathematical optimization \cite{kobler2017variational} or physical simulation \cite{meister2018towards} into deep learning pipelines.

\begin{figure}
    \centering
    \includegraphics[width=0.5\linewidth]{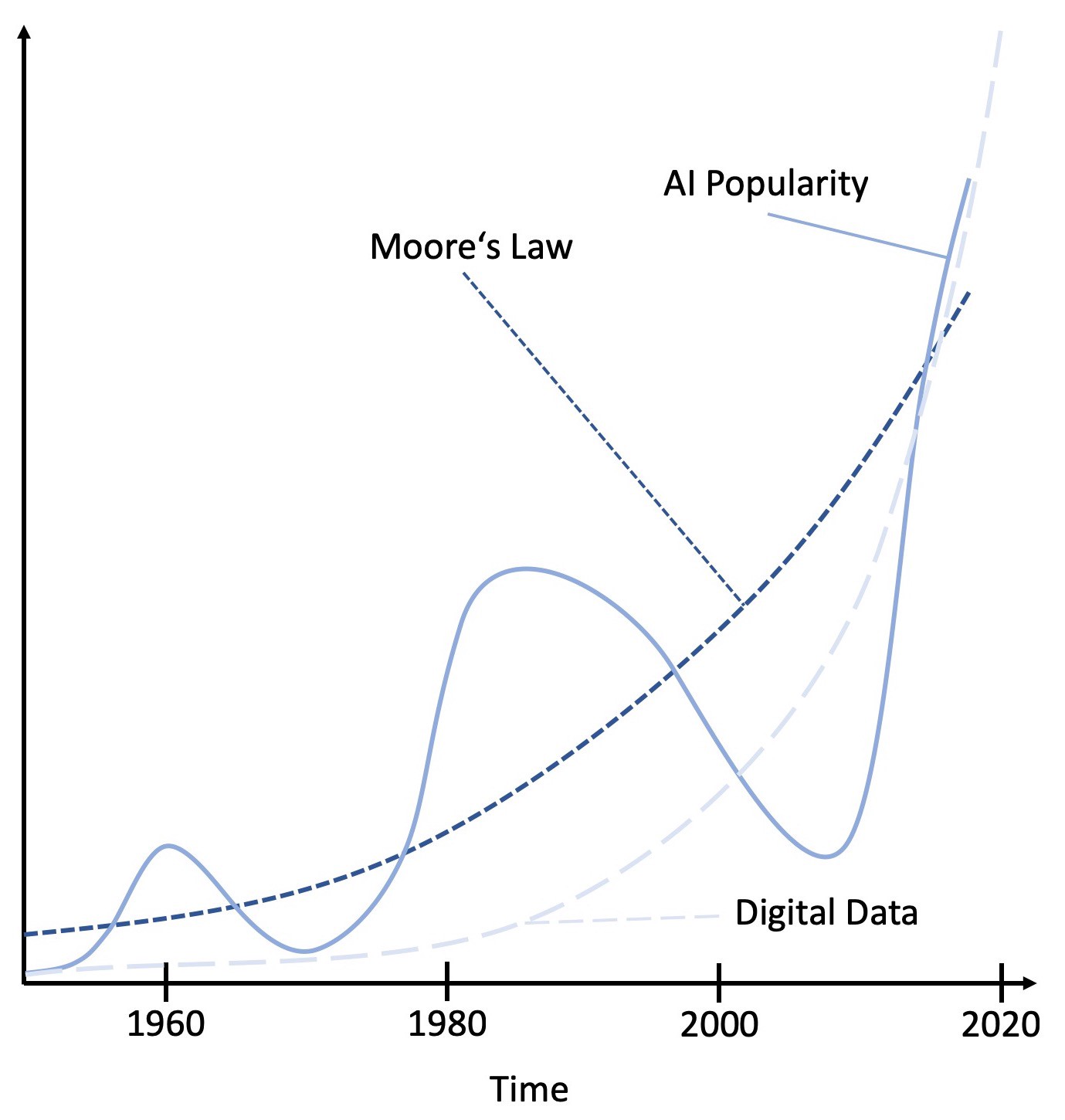}
    \caption{Both compute power and digital data grow exponentially. Yet, the high demands with respect to both cause regular impediments to the progress of general methods of artificial intelligence (AI).}
    \label{fig:ai_cycles}
\end{figure}

While these approaches show considerable success in terms of reduction of training data and generalization \cite{maier2019learning}, the idea of hybrid models has also been criticized by experts in the field of machine learning \cite{sutton2019bitter}. The line of argument is, that machine learning and artificial intelligence (AI) are generally compute- and data-bounded which causes the field to stall as soon as the respective resource is insufficient to guarantee the next successful step. As a result, the progress of the field runs in ``cycles'' as shown in Figure~\ref{fig:ai_cycles}. In order to compensate the lack of compute power, engineers would tend to mix known theory and heuristics with machine learning to create progress in absence of sufficient compute and data resources. The ``bitter lesson'' to be learned here is that as soon as enough compute power becomes available, general methods of AI surpass the hybrid approaches rendering them obsolete. As a result, Sutton \cite{sutton2019bitter} demands to stop all research on hybrids and to focus on general methods and in particular {\it meta learning} as this direction is likely to be the next major breakthrough in machine learning and AI.

Above argument is well founded and makes sense from a pure machine learning perspective. Yet, ``the purpose of computing is insight'' as already pinpointed by Hamming \cite{hamming2012numerical}. Hence, from a perspective of medical image processing, the creation of hybrid approaches makes in fact sense, as it allows to merge and actually enrich classical theory with modern-day machine learning ideas. Of course, many of the models that were identified so far in medical imaging \cite{maier2018medical} are approximations and do have shortcomings when implemented in real systems, following the famous words of Box ``All models are wrong, but some are useful.'' \cite{box1976science}. Yet, new experimental and theoretic evidence supports the use of such {\it prior knowledge}, {\it known operators} or {\it hybrid machine learning systems} now and also in the future \cite{hart2000pattern,maier2019learning}. 

For these reasons, the authors of this article believe that it is timely to perform proper review of the state-of-the-art in machine learning and medical image analysis regarding theory and applications of such hybrids. Note that in this article, we particularly focus on hybrids in contrast to other reviews that already address the general techniques on how to combine, e.g. physics with machine learning \cite{willard2020integrating}. 

In the following, we will describe the past with a particular focus on insights in machine learning which are very well in line with today's experimental findings. These theoretic observations are very well known in the field of machine learning, yet we believe that it makes sense to shortly summarize them in order to be able to follow present-day and possibly also future developments. The perspective of the past, will be followed by a review of the most visible grey box approaches in present-day medical imaging from medical image acquisition all the way to medical image analysis. The last part of the paper will be dedicated to future directions in the field and noteworthy approaches in machine learning that have a high potential to become relevant for the medical domain in the next months and years to come.

\section{The Past}

In the past, machine learning methods mainly focused on sub-tasks in medical imaging. In particular, in image segmentation they have been very successul \cite{zheng2014marginal}. Often machine learning is blended into longer algorithmic pipelines to solve specific challenges during the processing \cite{mualla2013automatic}. Therefore, much machine learning in medical imaging is a blend between both fields and there is no extended theory on how to perform such blends best. Therefore, a detailed look into the past of machine learning is advisable.

With the introduction of learning machines that were inspired by biological brains such as Rosenblatt's Percepton \cite{rosenblatt1960perceptron}, great interest in these technologies arose as already shown in Figure~\ref{fig:ai_cycles}. However due to limitations in compute-power and other theoretical doubts that were raised regarding the perceptron, the general interest declined again in later years. In fact several of such cycles were sparked, yet going through all of them in a historical perspective would well go beyond the scope of this article. The interested reader can find such historical perspectives published in  general media \cite{maier2019bittersweet, Aggarwal2018birth}. 

Instead, we will use this section to review three important theoretical results that will help us to understand the shortcomings in present-day machine learning for medical image processing better and will be relevant for our later analysis. The tree principles that we will shortly describe are the Universal Approximation Theorem which gives a fundamental argument why artificial neural networks work at all, the no free lunch theorem which tells us that there is no general way of determining the ``best'' algorithm, if we do not where it will be applied, and lastly the bias and variance trade-off which details that there is always a compromise to be made between generality and specialization of a learning system. 

\subsection{Universal Approximation Theorem}

The Universal Approximation Theorem \cite{cybenko1989approximation} is an important theoretical finding regarding which problems can be learned at all by an automatic system. In particular, it details that a single hidden layer artificial neural network $\classifier(\fex)$ is able to learn any continuous function $f(\fex)$ on a compact set, i.e.~a representative sample of the data at hand. Specifically, this can be written in the following way
\begin{equation}
    \classifier(\fex) = \sum_{i=0}^{N-1} v_i \activation({\bm w}_i^\top \fex +w_{0,i})
\end{equation}
where $N$ is the number of neurons, $v_i$ are weights of a linear combination of the hidden neurons, $\activation(\cdot)$ is the non-linearity or activation, and ${\bm w}_i$ and $w_{0,i}$ are the parameters of each neuron. Given such a constellation, an upper limit for the error $\epsilon$ can be found
\begin{equation}
    |f(\fex)-\classifier(\fex)| < \epsilon,
\end{equation}
which is valid over the entire domain of $\fex$. Closer analysis \cite{barron1994approximation} demonstrates that the error $\epsilon$ is dependent on the number of neurons $N$ and the amount of high frequencies in $f(\fex)$. With $N \rightarrow \infty$, $\epsilon$ approaches 0, i.e. technically any single layer network is able to approximate any function given a sufficient number of neurons.

These theoretical considerations seem to pose a contradiction to the observations in deep learning which could successfully demonstrate that multi-layer networks are superior to single layer networks in most applications \cite{lecun2015deep}. In contrast, theoretical analysis of error bounds \cite{barron2018approximation} found that the maximal approximation error is proportional to $L^\frac{3}{2}$, i.e. the more layers $L$ a net has, the higher the maximal error bound will be. So far, this apparent contradiction between theory and practice has not been resolved sufficiently. Still, the success in so many applications clearly demonstrates the power of deep networks \cite{crego2016systran, singh2017learning, yang2018introducing, christlein2018encoding, yang2019self}. 

\subsection{No Free Lunch Theorem}

The previous observations raise doubts given the apparent conflict between theory and practice. Yet, the Universal Approximation Theorem is not the only fundamental observation in the theory of machine learning. Therefore, we would like to revisit another important consideration which may help us to understand the previously mentioned conflicts: The ``no free lunch theorem'' \cite{hart2000pattern}.

The main take-away message of the theorem is that there is generally no ``best'' classifier, if the out-of-distribution error $\mathcal{E}_i$ for Algorithm $i$ on arbitrary functions $f(\fex)$ that are uniformly sampled from all possible functions $\mathcal{F}$, are considered. For two algorithms $\mathcal{A}_1$ and $\mathcal{A}_2$, the following observation is found
\begin{equation}
    \sum_f \left[\mathcal{E}_1(E|f, \mathcal{D}) - \mathcal{E}_2(E|f, \mathcal{D}) \right]= 0
\end{equation}
with $E$ being the expected miss-classification loss and $\mathcal{D}$ a fixed distribution of training samples. As such no difference in performance can be observed if all possible functions $f(\fex)$ are considered. Superiority of \textbf{any} classification algorithm can only be claimed if additional knowledge on the distribution of $f$, i.e.~the target application is known \cite{hart2000pattern}. This has a large effect on model selection and the success of machine learning models in practice and is referred to as \emph{inductive bias} in literature. 

As a result, particularly successful machine learning models encode knowledge about their application in order to outperform other approaches quite frequently. Prior knowledge, however, is not always available in many applications, a common approach to find optimal models is trial and error which led to the concept of ``graduate student descent'' \cite{gencoglu2019hark}. Some researchers also refer to such procedures as ``alchemy'' \cite{hutson2018has}.

\subsection{Bias and Variance Trade-off}

A last important theoretic observation of the past, that we would like to highlight here, is the ``bias and variance trade-off'' \cite{hart2000pattern} for general regression problems. It stems from the observation that the expected value $\avgerror$ of the square error on sampled data $\mathcal{D}$ can be decomposed in the following form:
\begin{equation}
    \avgerror_\mathcal{D} \left\{ (\classifier(\fex) - f(\fex))^2 \right\} =
    \underbrace{\avgerror_\mathcal{D} \left\{ \classifier(\fex) - f(\fex) \right\}^2}_{\textrm{Bias}^2} + 
    \underbrace{\avgerror_\mathcal{D} \left\{ \classifier(\fex) - \avgerror_\mathcal{D} \left\{ \classifier(\fex) \right\}^2\right\}}_\textrm{Variance}
\end{equation}

Methods with high bias are on average poor predictors of the function to be estimated, but generalize well to unseen data. Methods with high variance, typically have a low bias but tend to perform worse on unseen data. Above relation tells us that we can compensate high bias on data $\mathcal{D}$ by the introduction of a large variance and vice versa. This leads to a trade-off that allows us to pick between high error models that are likely to generalize or low error models that are less likely to work on unseen data. In any case, there is no general machine learning concept available that would allow to minimize bias \textbf{and} variance at the same time. The only method known to reduce both is the incorporation of prior knowledge on the target function $f(\fex)$ \cite{hart2000pattern}.

\begin{figure}
    \centering
    \includegraphics[width=1\linewidth]{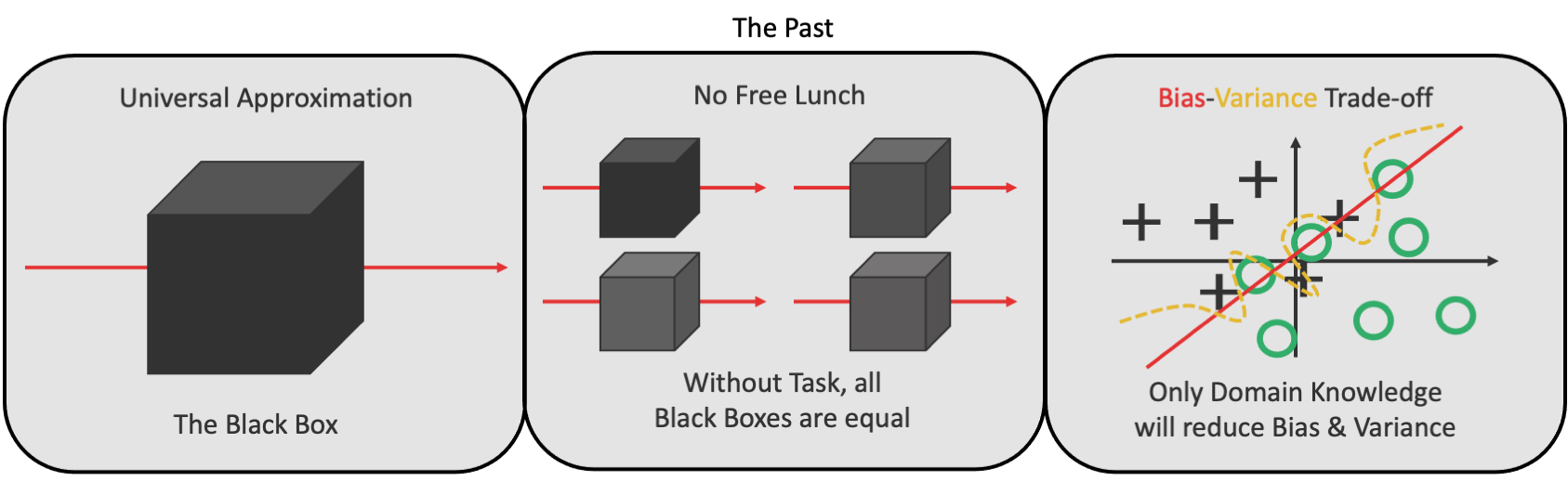}
    \caption{Major insights of past theory in machine learning regarding hybrid models can be summarized as follows: Universal approximation predicts that any function can be learned on a compact set. Without task, the no free lunch theorem predicts that all black box models are equal. The only path towards reducing bias and variance of any approach is the introduction of prior knowledge or \emph{inductive bias}.}
    \label{fig:thepast}
\end{figure}

All three observations that we just discussed highlight the importance of \emph{inductive bias} or prior knowledge for good performance of machine learning models (cf. Figure~\ref{fig:thepast}). Yet, past theory could only show its relevance, but did not develop generic concepts how such information could be encoded into machine learning models. Obviously a key challenge at the time was the limited amount of compute power that was available which essentially inhibited the fusion of complex compute routines describing prior knowledge and past machine learning approaches. This, however, changed at present day.

\section{The Present}

Today, we see a steady rise of deep learning models. Obviously, this paper is not able to summarize the entirety of deep learning methods. Even focusing only on the domain of medical image analysis would be way beyond the scope of a single paper. Yet, quite successful attempts of doing so are found in literature \cite{ker2018deep, litjens2017survey}.

In this section, we will shortly summarize important developments in the field and their relation to hybrid models. We first start looking from a high-level perspective onto the field of deep learning and highlight how different approaches manage to introduce \emph{inductive bias}. A competing approach as already mentioned by Sutton is obviously meta learning. We therefore have a quick look at the literature and summarize relevant methods that are being researched on today. Lastly, we will have a look at methods that implicitly or explicitly introduce prior knowledge into deep networks and summarized their strengths and weaknesses.

The perspective on theory will be followed by a short review of the state-of-the-art regarding hybrid approaches in medical image analysis featuring several highlight papers that the authors of this paper found particularly noteworthy.

\subsection{Theoretical Considerations}

\subsubsection{Deep Learning}

 is generally considered as artificial neural networks with many layers \cite{lecun2015deep}. Yet, interpretations of ``many'' vary broadly in literature from ``more than three'' up to thousands of layers. The network that caused a major breakthrough in image classification was Alex-net \cite{krizhevsky2012imagenet} which was able to cut the error on the image net challenge almost into half. In particular, the introduction of specialized operations into networks such as convolutional layers seem to be a key factor of the success of deep learning. It seems that the in-variances that are introduced by such layers are beneficial in terms of parameter reduction as well as with respect to incorporating \emph{inductive bias} into networks.

While convolutions are probably, the most popular layer that is used in deep learning, may other in-variances can also be encoded in network layers. Bronstein et al. summarized several important in-variances in a recent publication \cite{bronstein2021geometric}. As they demonstrate, grids, groups, graphs, geodesics, and gauges offer suitable in-variances which result in particular layer types ranging from convolutions over graph layers up to recurrent layers. Each one is able to describe a particular data in-variance that can be exploited in a particular mathematical operation. Yet, there are many more layer types which make use of such in-variances such as tensor networks which also allow the encoding of in-variances that correspond to higher-order tensor products \cite{biamonte2017tensor}. Even such techniques are already employed in in medical imaging, e.g.~image segmentation~\cite{selvan2020tensor}.

One question that cannot be answered sufficiently in literature at present is which in-variances to use and in which particular sequence and configuration. One key method to address the problem seems to be ``graduate student descent''.

\subsubsection{Meta Learning}

tries to learn the \emph{inductive bias} directly from the data itself. The general approach aims at learning similarities across different task domains. While a full summary of meta learning methods would again be beyond the scope of the paper, we will only describe few highlights in the past years in this section.

A model-agnostic approach for meta learning is presented in \cite{pmlr-v70-finn17a}. Here, the authors propose to disentangle the parameters of the meta task from the task-specific ones. This is achieved by modelling two sets of parameters: one for the generic task and one for the specific task. The devised training strategy produces stable parameters for the meta task and only few gradient iterations with few samples allow to adopt to a specific application.

In literature, several other approaches are found ranging from learning in-variances \cite{zhou2021metalearning} over learning prototypical networks \cite{snell2017prototypical} to neural architecture search \cite{zoph2017neural}. Yet, differences between the methods are small and vary from data set to data set. 

Unfortunately, a recent study suggests simply learning the task-dependent non-linear distance in feature space \cite{Sung_2018_CVPR} could demonstrate to be as effective as above mentioned meta learning approaches without even modelling the meta task. The paper demonstrates that common architectures which are used in deep learning are sufficient to fulfill this task.

It is clear that meta learning has high potential for future applications to model \emph{inductive bias}. Yet, none of the methods found in literature as able to demonstrate this for practical applications while outperforming today's deep learning methods.

\subsubsection{Prior Knowledge as Regularization} is another approach which attempts to enforce prior knowledge into deep learning models. The most frequent approach is to use the loss function to embed the prior knowledge using a regularization term \cite{willard2020integrating}. As such the loss is expanded by an additive part which punishes the model if it deviates from a priori knowledge given as an analytical model during the training. One major disadvantage of this approach is that the deviation from the physical model is only punished during training and there is no mechanism that actually guarantees the model's plausibility during test time. 

Another very common approach to force deep learning models to act plausible is to embed them into a constrained learning framework. This is often done by reinforcement learning or imitation learning. This approach was found to be very effective in playing games as the rules of the game can be enforced and sophisticated search algorithms such as Monte Carlo Tree Search \cite{silver2016mastering} can be embedded into the machine learning algorithms. This approach is quite popular also for applications in medical imaging as we will see in the application part \cite{ghesu2017multi}. Unfortunately, the reinforcement learning setup comes at rather high computational cost at training time and many episodes of the task at hand have to be repeated in order to train a reasonable system.

\subsubsection{Known Operator Learning} is an approach to directly embed analytical models into deep networks which are already known from classical theory. It is based on the assumption that the function to be learned can be decomposed into modules of which some are known and others have to be learned from data. Interestingly, there is also evidence that deep networks indeed frequently form such modular structures \cite{filan2020pruned}.

A key observation is that with every known operation that is introduced into the network, the maximum error bound is reduced \cite{maier2019learning}. A side effect of this procedure is that also the number of trainable parameters is reduced which empirically results in a reduced necessity for training samples. Also, the approach results in strong generalization capabilities such have also been demonstrated empirically \cite{syben2018deriving}.

Even without knowledge of above mentioned theoretical considerations, such hybrid models are frequently used in many task domains. In particular physics has a wealth of analytical models that are well suited for this approach. A recent paper \cite{li2021kohn} demonstrated that learning the entire hydrogen dissociation curve is possible and experimentally validated the generalization properties as predicted by known operator learning theory. Also the approach is suited to extract symbolic equations using graph networks as demonstrated in \cite{cranmer2019learning}.

In particular in the field of computer vision, the use of prior transforms and operations are becoming more and more popular. This ranges from simple geometric transforms such as spatial transformers \cite{NIPS2015_33ceb07b} up to complete differentiable rendering engines such as Rendernet \cite{nguyen2018rendernet}. A summary of the state-of-the-art in differentiable rendering approaches is readily found in the literature and would go beyond the scope of this perspective \cite{tewari2020state}.

Beyond differentiable rendering numerous other hybrid approaches are found in literature, such as relative pose estimation using deep networks encoding geometric transformations \cite{yang2020extreme}, deep point cloud rendering through multi-plane projections \cite{dai2020neural} or even direct rendering of voxels to images \cite{rematas2020neural}. Also multi-task approaches uniting deep depth, deep pose, and deep uncertainty estimation for monocular visual odometry \cite{yang2020d3vo} emerge on top of the use of hybrid models.

Not only physics and computer vision make use of such hybrid approaches. They already became popular and commonly used in all domains of signal processing. Applications in speech processing range from trainable filter-banks \cite{tukuljac2019spectrobank} to 
the integration of entire algorithms such as complex linear coding \cite{schroter2020clcnet}. 
There is even a software library for differentiable signal processing \cite{engel2020ddsp} available. 

\begin{figure}
    \centering
    \includegraphics[width=1\linewidth]{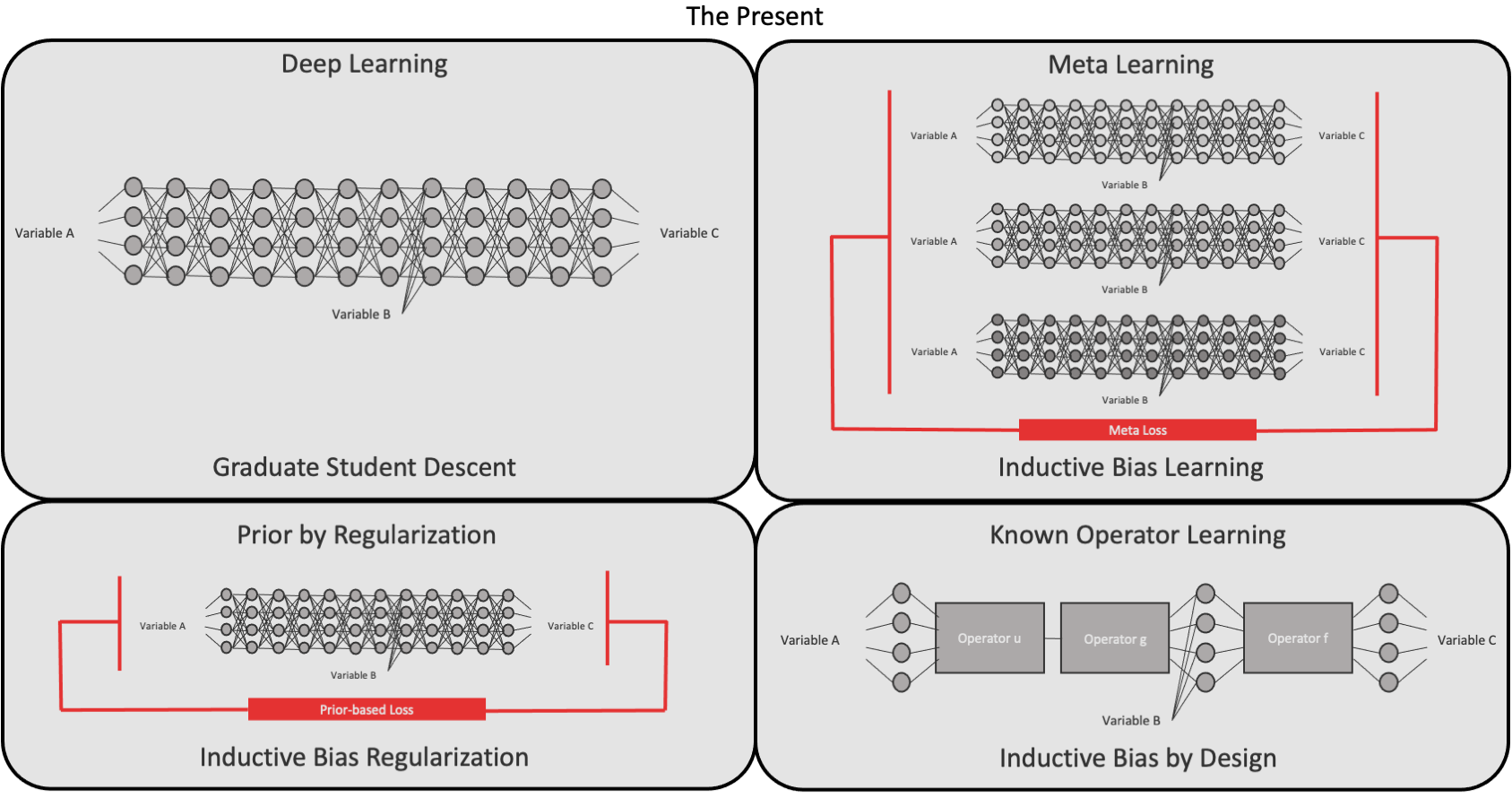}
    \caption{Several methods are common today to introduce \emph{inductive bias} into machine learning models. They range from ``graduate student descent'' in deep learning over meta learning to regularization methods. This article focuses on hybrid methods by design.}
    \label{fig:thepresent}
\end{figure}

Figure~\ref{fig:thepresent} summarizes today's approaches to introduce \emph{inductive bias} into machine learning models. All of the four presented approaches are successfully applied. In the near future, the methods which employ hybrids by design seem to be most promising ones to balance prior knowledge and modern-day deep learning approaches. In the following section, we will explore therefore current attempts to create such ``by design'' models in the medical imaging literature.

\subsection{Applications in Medical Imaging}

Also in the field of medical imaging and signal processing, hybrid methods are on the rise. It affects all domains from image acquisition to its analysis. In the following, we will discuss several highlights and across the entire domain.

\subsubsection{Scanner and Acquisition Design}

The first part of the whole image analysis chain is the actual image formation. It has numerous parameters which dramatically affect the image quality and subsequent image analysis. In particular, hybrid methods are well suited in this context as they allow linking of actual system parameters to the learning and analysis problem at hand.

In interventional applications of computed tomography, the choice of the scan trajectory, i.e. the sequence of X-ray source and detector positions is crucial and has large influence on the image quality. In a recent paper, Zaech et al. propose to formulate a learnable image domain task in order to determine an optimal scan trajectory for the particular analysis task at hand \cite{10.1007/978-3-030-32254-0_2}. In a follow-up application, Thies et al. expand the approach to effectively avoid metal artifacts during scanning during optimal learning-based trajectory design \cite{faucris.242204248}.

Also in the field of magnetic resonance imaging, optimal k-space sampling can lead to a reduction of scan time while preserving image quality. Pineda et al. suggest to do so using reinforcement \cite{pineda}. While the method is able to automatically design new scan patterns, it comes at the cost of an expensive episode-based learning program. In contrast, Zaiss et al. demonstrated that actually the entire k-space acquisition process is differentiable \cite{mrzero}. As a result, sampling patterns and contrasts can be learned directly using the efficient methods of supervised learning yielding the possibility to mimic arbitrary contrasts from other modalities such as PET and SPECT.

\subsubsection{Image Reconstruction}

Closely connected to the actual acquisition is the image reconstruction which actually forms the final image to be analysed. Here, complete black box approaches have been suggested \cite{zhu2018image} which come at the great risk to ignore all task-domain knowledge about imaging physics. While published results look impressive, the resulting images also come at the risk that pathologies might be hallucinated into the image or --- in an even worse scenario --- might be missing entirely \cite{huang2018considerations}.

Therefore, also in this field, hybrid methods are on the rise. First approaches started with mapping classical filtered back-projection algorithms onto neural networks using differentiable back-projection \cite{wurfl2016deep, wurfl2018deep}. Kobler et al. demonstrated that unrolling techniques which are commonly used with recurrent neural networks enable the mapping of iterative reconstruction approaches to trainable reconstruction algorithms \cite{kobler2017variational}. Most noteably, this was done by Hammernik et al. for MRI \cite{hammernik2018learning} and CT \cite{hammernik2017deep}. Later, Adler et al. demonstrated that also more advanced methods such as primal-dual reconstruction can be mapped onto deep networks \cite{adler2018learned}. Today, most commonly CNNs are used create learned regularizers \cite{li2020nett}. Furthermore, data
consistency layers are used frequently which effectively make sure that only data is interpolated within the null space of the reconstruction
operator \cite{schwab2019deep}. This procedure guarantees that the actually observed data is preserved in the reconstruction process while only unobservable data points are interpolated by the deep learning approach. Based on these results, it is clear today that virtually all classical energy minimization-based methods can be mapped onto trainable variants and therewith be made ``deep learning compatible''.

Above theoretical considerations has lead to a multitude of applied reconstruction results such as mapping of MRI image reconstruction onto ordinary differentiable equations and embedding them into a deep network \cite{chen2020mri}, deep cascades of convolutional neural networks for dynamic MR image reconstruction \cite{Schlemper2018}, U-net cascades with data consistency layers \cite{kofler2018u}, and variational networks in ultrasound \cite{vishnevskiy2018image}.

Ye et al. demonstrated a surprising analogy between so-called Generative Adversarial Networks (GANs) and the forward and inverse models used in image reconstruction \cite{zhu2017unpaired}. This particular setup follows the principle of a cycle GAN and therewith also allows unpaired training approaches while embedding task-specific domain models. This was successfully demonstrated for microscopy deconvolution \cite{lim2020cyclegan} and MRI acceleration \cite{oh2020unpaired}.

In a more ad-hoc manner, also other \emph{inductive biases} can be implemented, e.g. Schirrmacher et al. showed how to integrate median and quantile image filters \cite{miccai:schirrmacher}, Maier et al. integrated heuristic convolutional scatter models \cite{maier2018deep} while Roser et al. preferred splines for this purpose \cite{9410231}. Even metal artifact reduction can be made trainable using hybrids as demonstrated by Gottschalk et al.  \cite{gottschalk2021learningbased}. It is foreseeable that many other classical physics-based modelling approaches will soon follow in this line of research.

\subsubsection{Image Segmentation} has seen dramatic progress since the introduction of deep learning-based approaches. In particular, U-net \cite{ronneberger2015u} became one of the most popular approaches. At it's core it integrated up- and down-sampling steps to integrate the multi-scale nature of the image segmentation process. Quite popular are extensions to 3-D such as V-net \cite{milletari2016v} and ideas to automatically determine the hyperparameters \cite{isensee2018nnu}.

The overwhelming dominance of U-net type of approaches is only scarcely rivaled. A very different and therefore noteworthy approach is to use recurrent neural networks for segmentation \cite{andermatt2016multi} in order to model spatial co-locations of different landmarks with respect to each other.

Lately, also other hybrid approaches emerged to counter negative side effects of pure deep learning based approaches. Based on the vesselness after Frangi et al. \cite{vesselness-frangi}, Wu et al. created a hybrid neural network \cite{ArXivWeilin} with comparable performance to U-net at a fraction of U-net's parameter. Later Wu et al. also showed that U-net type of networks are able to learn optimal pre-processing networks for the vesselness filter \cite{10.1007/978-3-030-32239-7_21} rendering them task-agnostic.

Further recent hybrids in image segmentation build on deep active contours \cite{zhang2020deep}, differentiable meshing \cite{wickramasinghe2020voxel2mesh}, and deep action learning \cite{faucris.250564935} for active shape models.

\subsubsection{Image Registration}

In contrast to image reconstruction, there are only few approaches in image registration which make use of explicit prior knowledge. Most techniques focus on the direct prediction of the deformation field given the moving and the target image such as Quicksilver \cite{yang2017quicksilver} or Voxelmorph \cite{8633930}.

Based on the ideas of reinforcement and imitation learning, also classical models can be employed, as shown by Liao at al. for rigid registration \cite{liao2017artificial} and Krebs et al. for the non-ridig case \cite{univis91731175}.

For 2-D/3-D registration Schaffert et al. demonstrated that the occulding contour technique can be embedded into a deep network as a known operator allowing to learn task-specific key point matching techniques \cite{schaffert2018metric}. Another noteworthy approach in this context was presented by Gao et al. \cite{10.1007/978-3-030-59716-0_32} 
as they generalize spatial transformers for 2-D/3-D registration of X-ray projections and CT volumes.

\subsubsection{Image Generation \& Simulation} are two hot topics in deep learning. For image generation, deep networks and GANs are particularly popular, e.g. as shown for MRI to CT image conversion \cite{han2017mr}. Yet, GANs are known to hallucinate features \cite{Cohen2018distribution} which is also observed in practice \cite{schiffers2018cyclegan}. To avoid such problems, it is adviseable to embed hybrid methods. Stimpel et al. demonstrate this for upsampling of MRI images using a trainable guided filter \cite{stimpel2019multi}. Also the generation of close to realistic X-ray images can benefit from hybrid approaches as presented in Deep DRR \cite{10.1007/978-3-030-00937-3_12}. Here, geometry and absorption are modelled using analytical tools while the stochastic anatomy, scatter, and noise are generated using deep learning-based ideas. 

A recent, but very promising approach embeds trainable cell automata to model the growth of tumors \cite{manzanera2021patient}.
While the study is preliminary, first impressive results are shown for realistic nodule generation.

As many deep learning methods require large amounts of training data, accurate physical rendering is an alternative to GANs in the creation of training data. Mill et al. use rendering to generate realistically looking images and  corresponding annotations \cite{mill2021}. In their results, they show for a variety of nano particles that this approach is en par to manual annotation of real images.

Also in physical simulation, hybrid methods are on the rise. This ranges from acceleration of classical simulation optimizers by learning gradient steps \cite{meister18:TFB} to the integration of entire physical simulation engines into the machine learning algorithm \cite{NEURIPS2020_43e4e6a6}. This approach is particularly interesting as it enables to blend methods of deep learning and classical simulation elegantly.

\section{The Future}

\begin{figure}
    \centering
    \includegraphics[width=1\linewidth]{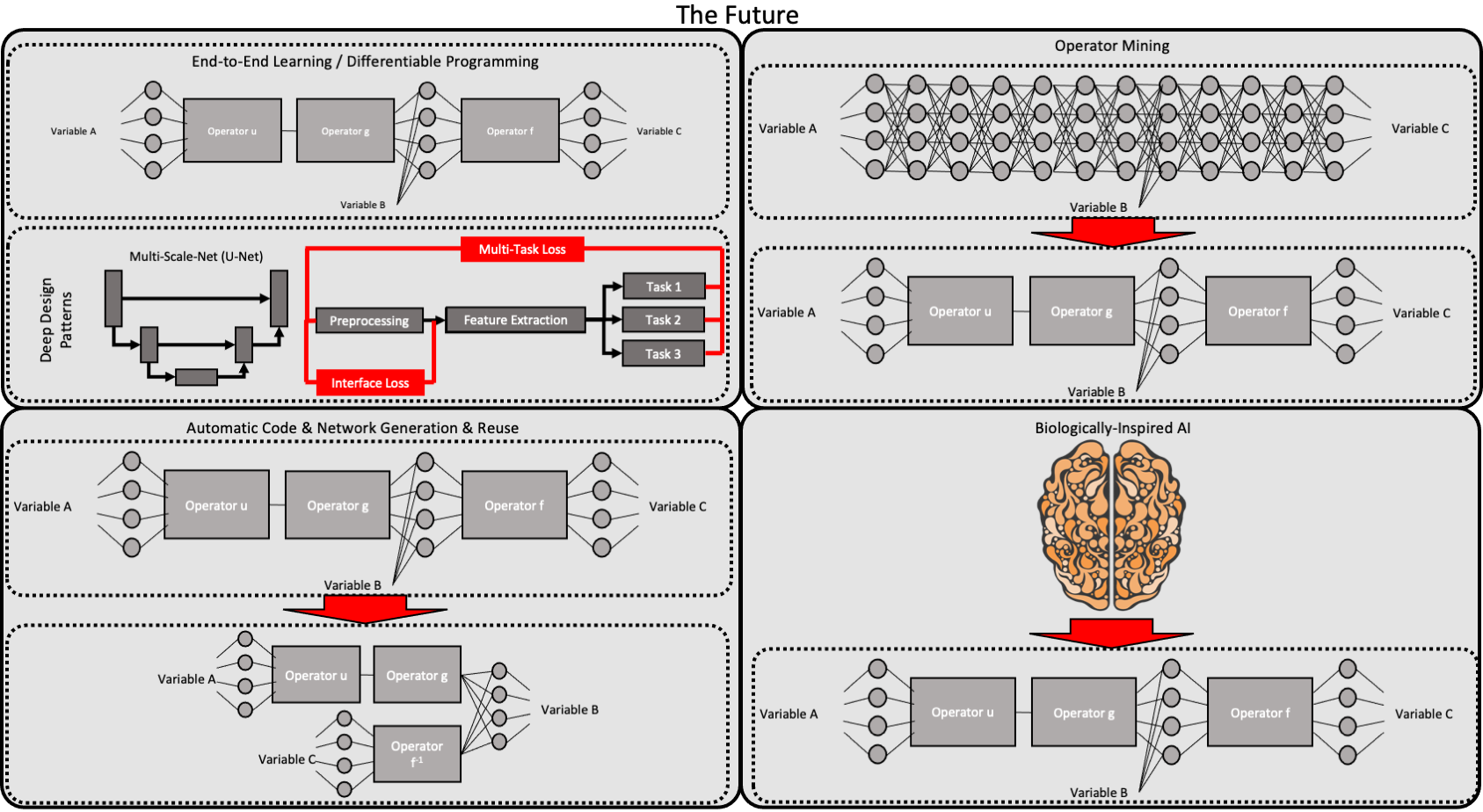}
    \caption{Based on the current approaches, it is foreseeable that several new technologies will emerge. Already a current trend are architecture blocks that are re-used in other settings. Such ``deep design patterns'' are likely to be applicable over a wide range of applications. With techniques from code reverse-engineering deep networks can be analysed similar to assembler and byte-code to obtain higher-level operations using ``operator mining''. Based on structured nets of known operators, tranformations on already trained networks can be performed. Doing so would allow to re-use them towards entirely new purposes based on the techniques of automatic code generation. Lastly, the human brain is an incredible resources which already solves many of the problems that we face in machine learning today. Therefore, also biologically-inspired AI will play a major role in machine learning models of the future.}
    \label{fig:thefuture}
\end{figure}

As we could see in the previous section, present-day methods already involve a multitude of hybrid machine learning models. Most of the results that we found in literature were in line with the theoretical predictions of known operator learning. Yet, Sutton is correct about the need to drive such developments automatically. However, looking at the true plurality of the hybrids in medical applications that we highlighted here, it becomes clear why present-day meta learning models frequently fail to incorporate \emph{inductive bias} sufficiently. For these reasons, we will look into several promising ideas which we will detail in the following. As a matter of fact, they address general problems in machine learning and are likely to be applicable to other task domains beyond medical image processing as well. A graphical summary of this section is found in Figure~\ref{fig:thefuture}.

\subsection{Differentiable Programming \& Deep Design Patterns}

Already with the development of programming environments such as Tensorflow \cite{abadi2016tensorflow} and PyTorch \cite{paszke2019pytorch}, it became clear that the age of deep learning will require new programming paradigms. The next step is likely to be differentiable programming \cite{li2018differentiable} in which all programming constructs are described in a way in which gradients for the trainable part of the software can be generated automatically. For specific task-domains such programming environments already exist, e.g. for signal processing 
\cite{engel2020ddsp} or medical image reconstruction \cite{syben2019pyro, ronchetti2020torchradon}.

Another direction which is likely to emerge are ``Deep Design Patterns''. Similar to the software design patterns \cite{gamma1993design}, we see also many patterns in the development of deep networks such as multi-task losses, U-Nets, and transformers. Yet, we still discover new patterns such as the interface pattern as introduced in \cite{fu2019lesson} which strategically chooses losses, e.g. to mimic certain intermediate results to enforce a data interface at a particular layer. As observed by Wu et al. this is also key to enable re-use of already trained networks.

\subsection{Network \& Operator Mining}

Deep networks share many similarities with compiled assembler or byte-code. The code can be run on native hardware, yet it is very difficult to interpret. Therefore, methods from disassembly \cite{pro2012interactive} or code lifting \cite{rohleder2019hands} seem to be well suited to identify design patterns and known operators in deep networks. Partially, we already see methods following very similar approaches such as the physical equation discovery via candidate model generation and search in \cite{simidjievski2020equation}.

\subsection{Automatic Code \& Network Generation and Reuse}

A recent paper follows the paradigm of known operator learning and is able to derive new deep network architectures \cite{syben2018deriving} driven by mathematical properties of the involved operations and desired features of the algorithm. In our view, this is of course only the first step. Given a trained network and the mathematical properties of its operators, the complete deep net can be reformulated mathematically to a new purpose, as shown in the bottom left of Figure~\ref{fig:thefuture}. For operations with known inverses, these can be employed directly. For any other operation, techniques from inverse problems will allow their re-wiring without the need for re-training or fine-tuning of the network. Of course, such reformulation cannot be done manually which will require automatic code transformation techniques.

\subsection{Biologically-inspired AI}

The human brain already solves many of the tasks that we try to address in machine learning and AI. In fact, we already use such insights in many occations in machine learning and image processing. For instance, the classical perceptron was inspired by the biological neuron \cite{rosenblatt1960perceptron}. In addition, convolutional layers \cite{lecun2015deep} resemble the localized connectivity patterns in the mammalian visual system \cite{kandel2000principles}. Here, the neurons' receptive fields correspond to the kernels of convolutional networks. Another example is the architecture of residual networks \cite{he2016deep} and U-nets \cite{ronneberger2015u} with numerous parallel layers at the same hierarchy level, and connections that bypass several hierarchical levels (the residual connections). These design principles are strikingly similar to the complex, parallel-hierarchical connectivity of areas in the primate cerebral cortex \cite{felleman1991distributed}. Finally, the idea of layer-wise training of stacked auto-encoders \cite{erhan2010does} or deep belief networks \cite{hinton2006fast} which is a solution to the the problem of exploding or vanishing gradients \cite{bengio1994learning}, also has its model in neurobiology: Due to the gradual, progressive myelination of nerve fibers between cortex areas, further areas are successively functionally connected during development, starting with the primarily sensory and motor areas \cite{kandel2000principles}. In this way, the number of connected cortex areas and thus the size of the cortical network grows during development. In humans, this process begins at birth and continues until the end of the third decade of life \cite{miller2012prolonged}.

To draw a conclusion from the above list of examples of biologically-inspired AI -- which is by no means exhaustive -- it is very likely that more discoveries are to be made by the analysis of structure and function of the human brain. For instance, noise-induced resonance phenomena such as stochastic \cite{benzi1982stochastic} or recurrence resonance \cite{krauss2019recurrence} are known to play a crucial role in biological neural networks \cite{moss2004stochastic, mcdonnell2009stochastic, krauss2018cross} where they enable robust and flexible information processing \cite{krauss2016stochastic, krauss2017adaptive, schilling2020intrinsic, krauss2021simulated}. A further example is the hippocampal formation which is known to solve general mapping problems and to enable flexible multi-scale routing through arbitrary cognitive and conceptual spaces \cite{stachenfeld2017hippocampus, bellmund2018navigating, momennejad2020learning, park2020map} and are likely useful to develop corresponding machine learning counterparts. Obviously, also the answer for human-level few-shot and meta learning, as well as for other shortcomings of contemporary machine learning pointed out by Gary Marcus \cite{marcus2018deep} lies in the brain \cite{krauss2020will}.

\section{Summary}

In this paper, we reviewed the past, the present and the future of hybrid machine learning systems with a particular focus on medical image processing. The past showed us that classical machine learning has no clear approach to tackle \emph{inductive bias}. Yet, it acknowledges its importance. For these reasons, we see a multitude of approaches to address this problem today ranging from ``graduate student descent'' over meta learning to direct inclusion of prior knowledge. In particular the ``by design models'' in the spirit of the known operator paradigm flourish in today's research. Given the vast amount of choices for such hybrids, it is clear why current meta learning approaches have trouble discovering all of them. Our analysis is very much in line with the predictions already made by Gary Marcus for the next decade of AI \cite{marcus2020decade}.

For the future, we see in particular deep design patterns, known operator mining, automatic network re-wiring \& re-use, and in particular biologically-inspired approaches becoming more and more relevant. Yet, as this task is complex, we will be able to easily spend the next decade of machine learning research to reach these goals.

\section*{Acknowledgments}
This work was funded by the Deutsche Forschungsgemeinschaft (DFG, German Research Foundation): grant KR5148/2-1 to PK -- project number 436456810, the Emerging Talents Initiative (ETI) of the University Erlangen-Nuremberg (grant 2019/2-Phil-01 to PK). Furthermore, the research leading to these results has received funding from the European Research Council (ERC) under the European Union's Horizon 2020 research and innovation programme (ERC grant no. 810316 to AM).

%\section*{Author contributions}
%Both authors contributed equally to this work.

%\section*{Competing interests}
%The authors declare no competing financial interests.

% *************************************************************
% REFERENCES
% *************************************************************

\section*{References}
\bibliographystyle{iopart-num}
\bibliography{refs}

\end{document}